# Creativity of Deep Learning: Conceptualization and Assessment

Marcus Basalla[1], Johannes Schneider[1] and Jan vom Brocke[1]
[1]*Institute of Information Systems, University of Liechtenstein, Vaduz, Liechtenstein*
*{marcus.basalla,johannes.schneider,jan.vombrocke}@uni.li*

Abstract: While the potential of deep learning(DL) for automating simple tasks is already well explored, recent research has started investigating the use of deep learning for creative design, both for complete artifact creation and supporting humans in the creation process. In this paper, we use insights from computational creativity to conceptualize and assess current applications of generative deep learning in creative domains identified in a literature review. We highlight parallels between current systems and different models of human creativity as well as their shortcomings. While deep learning yields results of high value, such as high-quality images, their novelty is typically limited due to multiple reasons such a being tied to a conceptual space defined by training data. Current DL methods also do not allow for changes in the internal problem representation, and they lack the capability to identify connections across highly different domains, both of which are seen as major drivers of human creativity.

# 1 INTRODUCTION

The year 2019 can be seen as the year when artificial intelligence(AI) made its public debut as a composer in classical music. On February 4th, Schubert's unfinished 8th Symphony was performed in London after being completed by an AI system developed by Huawei (Davis, 2019). Later in April, the German Telekom announced their work on an AI to finish Beethoven's 10th Symphony for a performance celebrating the 250 years since the birth of the famous German composer (Roberts, 2019). While the quality of the AI's composition has been under scrutiny (Richter, 2019), it is nevertheless remarkable and resulted in the public and corporations' large interest in using AI for such creative fields.

For a long time, creating and appreciating art was believed to be unique to humans. However, advancements in the field of computational creativity and increased use of artificial intelligence in creative domains call this belief into question. At the core of the current rise of AI is deep learning (DL), fuelled by increasing processing power and data availability. Deep learning went quickly beyond outperforming previous solutions on established machine learning tasks to enable the automation of tasks that could previously only be performed with high-quality outcomes by humans, like image captioning (You et al., 2016), speech recognition (Amodei et al., 2016), and end-to-end translation (Johnson et al., 2016). At the same time, advanced generative models were developed to generate images and sequences of text, speech, and music. Such models proved to be a powerful tool for creative domains like digital painting, text- and music generation, e.g., AI-generated paintings have been sold for almost half a million USD (Cohn et al., 2018). But while these examples point to a high potential of DL in creative domains, there is so far no comprehensive analysis of the extent of its creative capabilities. A better understanding of the creative capabilities of deep learning is not only of general public interest, but it helps improve current generative DL systems towards more inherent creativity. It also helps companies better assess the suitability of adopting the technology. For example, it could be beneficial to integrate deep learning technology into creative human workflows, e.g., to provide suggestions for improvement to humans (Schneider, 2020). As any technology can be abused as well, an understanding of the creative potential is also relevant to anticipate and protect against malicious intent, e.g., in the form of deception (Schneider, Meske et al., 2022). We, therefore, pose the following research question:

*To what extent does deep learning exhibit elementary concepts of human creativity?*

To shed light on this question, we derive a conceptualization of human creativity based on computational creativity works and conduct a literature review on creative AI and applications of deep learning models. We also assess these works concerning creativity according to our conceptualization.

We observe that generative DL models mimic several processes of human creativity on an abstract level. However, the architecture of these models restricts the extent of creativity far beyond that of a human. Their creative output is also heavily constrained by the data used to train the model resulting in relatively low novelty and diversity compared to the data. Furthermore, while in some domains creative solutions are of high value, e.g., generated images are of high quality, in other domains that require multiple sequential reasoning steps, they are limited in value, e.g., in storytelling, where they fail to capture a consistent theme across longer time periods.

# 2 METHODOLOGY

We first derive a conceptualization of human creativity consisting of 6 dimensions, based on established concepts from the computational creativity domain (Wiggins et al., 2006; Boden et al., 1998). The concepts are rooted in human creativity. Therefore, they are not limited to a specific family of AI algorithms. They allow us to draw analogies to humans more easily. We conduct a qualitative literature review (Schryen et al., 2015) that focuses on creative AI and applications of DL models. We use the literature to refine our conceptualization and to build a concept matrix (Webster & Watson, 2002). We then support the validity of our framework by showing parallels with other theories of human creativity and investigating how DL ranks on each dimension of creativity of our conceptualization.

For the literature review, we performed a keyword search on the dblp computer science bibliography (Ley et al., 2009), focusing on articles published in journals and conference proceedings. To capture an extensive overview of the literature on computational creativity the keywords "computational creativity" alongside combinations of the keywords "creativity AND (AI OR Artificial Intelligence OR ML or Machine Learning OR DL OR deep learning OR Neural Network)" were used. We limited our search to papers after 2010 as this was at the offset of the rise of deep learning (Goodfellow et al., 2016). From these, all papers that describe a creative design process that applied DL were manually selected. This left us with a list of 18 papers. It was enhanced through forward- and backward searches based on the 18 identified papers. All in all, this process left us with a selection of 34 papers describing generative applications of DL.

# 3 FRAMEWORK

## 3.1 Creativity

Boden et al. (1998) define a creative idea as "one which is novel, surprising, and valuable". The two key requirements for creativity, novelty and value, are found in one way or another in most definitions of creativity. Thus, we define creativity as a process that generates an artifact that is both novel and valuable. In other words, creative artifacts must differ from previous artifacts in the same domain (novelty) while still fulfilling the purpose they were intended for (value). A random combination of shapes and colors in itself is, for example, not a creative piece of art, if the art's purpose is to show an abstraction of an actual object or to elicit an emotional or aesthetic response in the observer. On the other hand, adding a few new lines to provide more details to an existing painting might change its aesthetic. However, it would hardly be considered novel.

One can further categorize creativity by their output as mini-c, little-c, pro-c, and Big-C creativity (Kaufman & Beghetto, 2009). Mini-c and little-c creativity are concerned with everyday creativity. Little-c creativity is concerned with creative processes that generate tangible outputs, whereas mini-c only requires a novel interpretation of certain stimuli like experiences or actions. Big-C creativity is concerned with creative outputs that have a considerable impact on a field and are often connected with the notion of genius. Pro-c creativity is concerned with outputs by professionals recognized as being novel to a domain but without revolutionizing or strongly influencing the domain.

## 3.2 Dimensions of Creativity

Boden et al. (1998) define three types of creativity: combinational, explorational, and transformational creativity. All three mechanisms operate in a conceptual space. This conceptual space can be interpreted as the cognitive representation or neural encoding of a person's understanding of the problem domain. Wiggins et al. (2006) further clarify the definition of a conceptual space by introducing a search space and a boundary. The boundary is a meta description that defines the boundary of possible search spaces. It contains all ideas of boundary definitions that a person can conceive of based on their problem or domain understanding. The search space defines all ideas that a creative person (or AI) can conceive of using a specific method of ideation. The search space is a subset of the conceptual space, while the boundary defines the extent of the conceptual space. For example, for playing chess, the boundary might be the number of rounds considered for the current board, e.g., player A moves a figure, player B moves a figure, etc. The search space would be the total number of moves. The left panel Figure 1 shows our model of creativity based on the aforementioned computational creativity works. The problem (understanding) informs the boundary of the conceptual space, limiting the extent of all possible search spaces. Generic methods of ideation on a specific search space, i.e., the forming of concepts and ideas, result in creative solutions to the problem.

## 3.3 Creativity models: A human and DL perspective

While the left panel in Figure 1 shows our model of creativity based on the computational creativity works, the right panel in Figure 1 shows a related model of creativity based on common concepts in machine learning. While we shall focus on computational creativity, since it is closer to a human notion of creativity, it is also insightful to derive a model of creativity inspired by machine learning.

While computational creativity might be said as moving from more abstract, broad, non-mathematically described human concepts of creativity towards a more concise computational perspective. The machine learning-based model might move from a mathematically well-defined, more narrow computational perspective of creativity towards human concepts. Therefore, the matching elements in both models, such as search space and parameters, are not identical. Parameters are typically a set of real numbers within a DL model optimized in the training process using a well-known method, e.g., stochastic gradient descent in DL. In contrast to this mathematically sound but narrow view, the search space in computational creativity is vaguer and broader. The same logic for distinction applies when comparing the boundary restricting and defining the search space and the meta-parameters defining the DL model (and its parameters). We discuss this in more detail, focusing on generative deep learning, which we view as a key technology for creativity within DL.

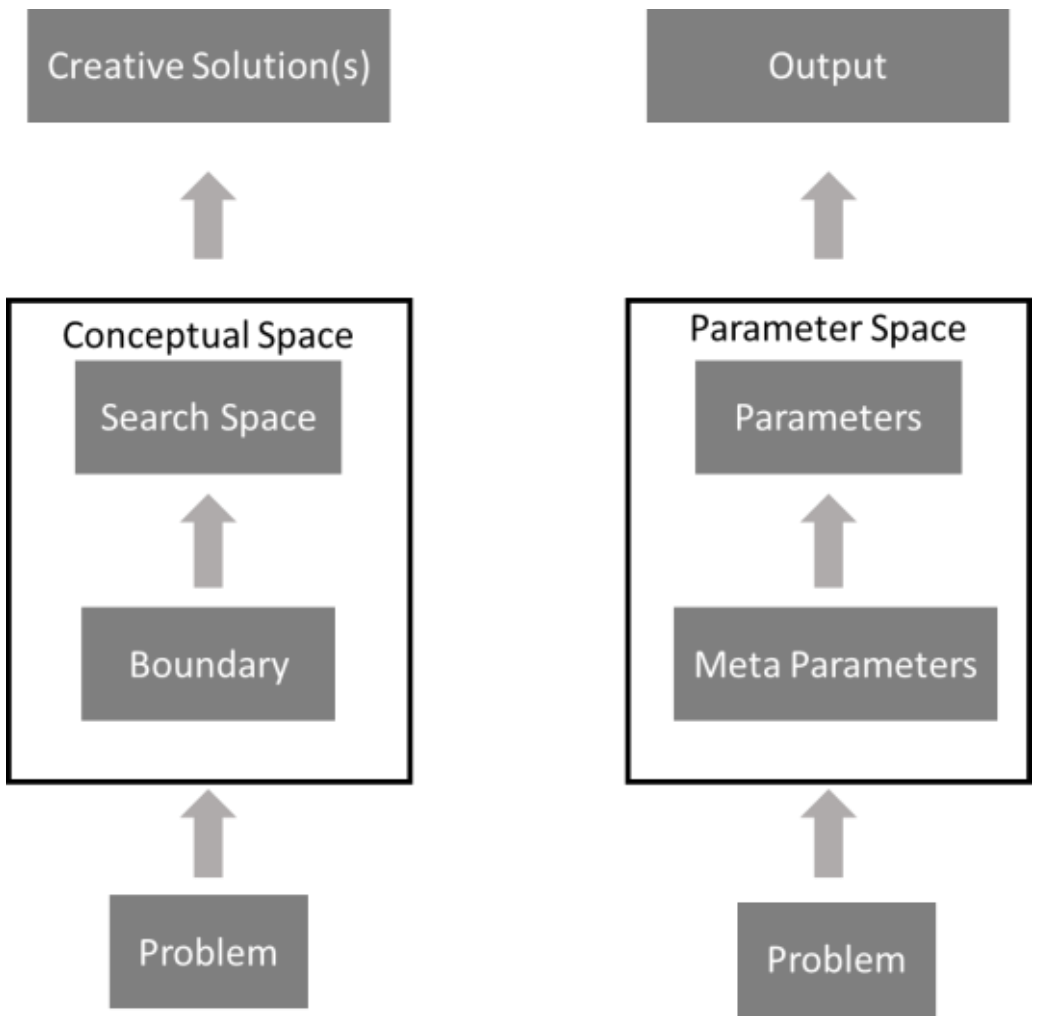

**Fig. 1.** Creative process model as found in computational creativity inspired from humans (left) and a model inspired from machine learning (right); recursive connections are not shown

## 3.4 Parallels to Generative Deep Learning

A key element of deep learning is representation learning (Bengio et al., 2013). Data is represented through a hierarchy of features, where each feature constitutes a frequent pattern in the data. Typically, for classification networks, layers closer to the input resemble simpler, less semantically meaningful samples than layers closer to the output. Generative deep learning networks are trained to approximate the underlying latent probability distribution of the training data with the learned representation. New outputs are generated by sampling from this distribution.

By drawing parallels between generative DL and our framework, it becomes evident that the problem representation, which is encoded by the network parameters, can be seen as an equivalent to the search space in the creativity framework by (Boden et al., 1998), where sampling from this distribution to generate new outputs can be seen as a process to generate new creative outputs.

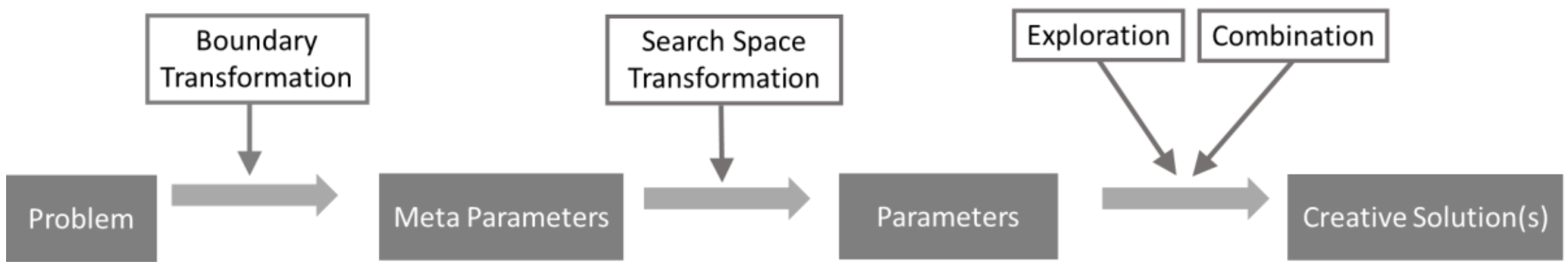

**Fig. 2.** Elements of the conceptualization expand the creative process model in Figure 1.

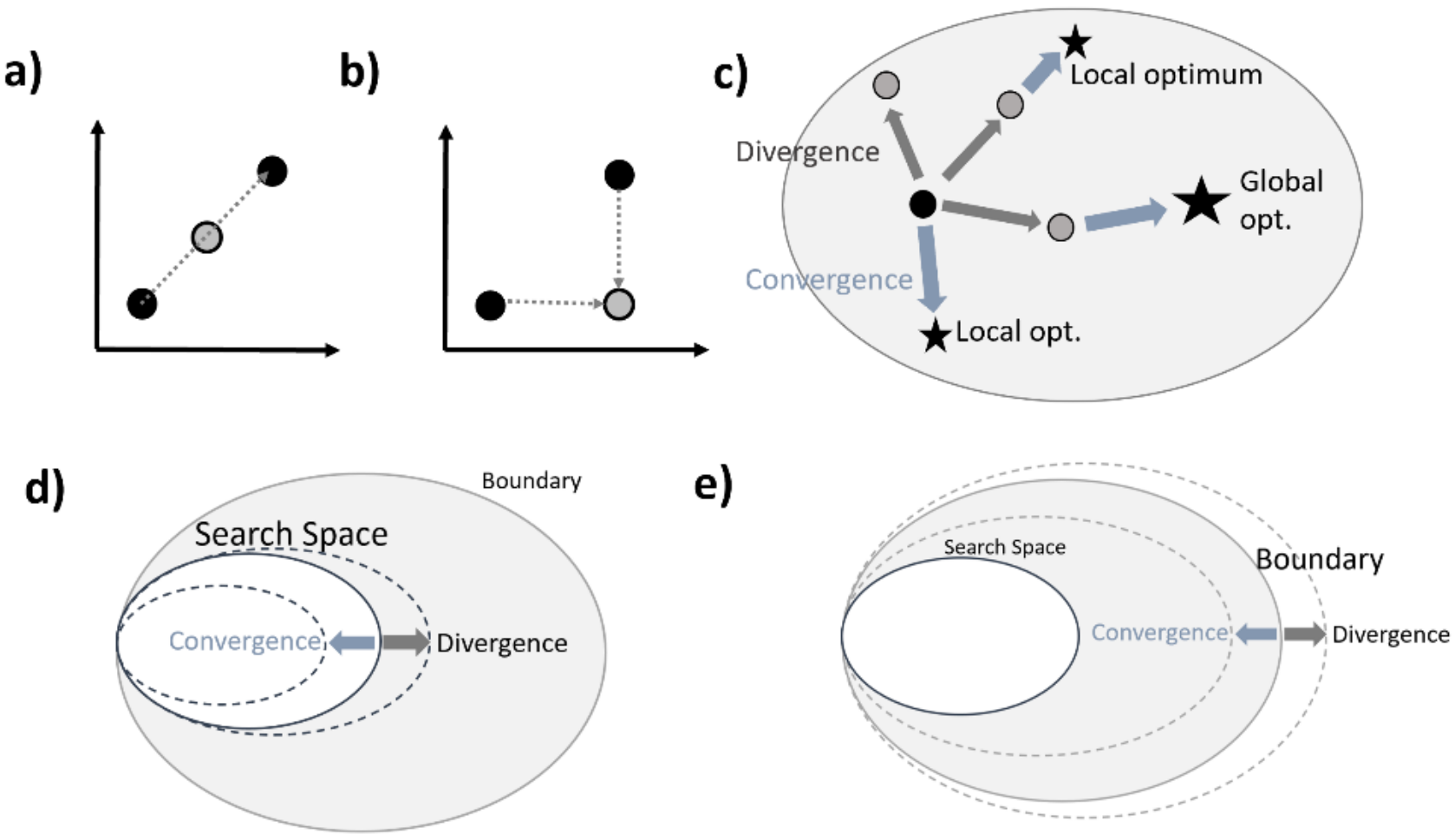

**Fig. 3.** (a) Interpolation and (b) Recombination in 2d feature space; Divergent and convergent (c) Exploration, (d) Search space transformation and (e) Boundary transformation.

We can use meta-learning to find an equivalent to the boundary (Hospedales et al., 2020). Meta-learning differentiates between the model parameters $\theta$ and meta knowledge $\omega$, which incorporates all factors that cannot directly be trained by training methods such as gradient descent, like the network architecture and models hyperparameters (Huisman et al., 2021). Meta-learning itself requires a concise, mathematical description, which limits the possible boundaries. Furthermore, this description originates from humans. The search space in Figure 1, from which solutions can be generated, relates to the network, i.e., its feature representation being equivalent to the fixed parameters of the model. The network features originate from the boundary using a training process and the training data. The network takes inputs and provides outputs. The search space corresponds to the set of all possible inputs that the network can process.

## 3.5 Creative Processes

Next, we introduce the dimensions, which describe the creative process and categorize existing works on DL for creative domains. A summary is shown in Figure 2.

**Exploration:** Explorational creativity describes the process of generating novel ideas by exploring a known search space. Solutions that are hard to access in a specific search space are generally more novel, especially considering the perspective of other creators that work in the same search space (Wiggins et al., 2006). Therefore, this category can include any search strategy if it does not manipulate the search space. In theory, the most creative solution might be found by investigating all possibilities, but this is computationally infeasible due to the size of the search space. A good strategy can narrow the search space to more novel and more valuable sub-spaces.

**Combination:** Combinational creativity describes the process of combining two or more known ideas to generate novel ideas. Ideas can be combined, if they share inherent conceptual structures or features (Boden et al., 1998). Low creativity is indicated by combining similar ideas. High creativity is indicated by combining diverse ideas (Ward & Kolomyts, 2010). As the specific combination process is left general, this can include several processes that interpolate between features (Figure 3a) or recombine features (Figure 3b) of known solutions. Identifying a solution using "analogies" is an example of combinational creativity (Ward & Kolomyts, 2010).

Combination and transformation are not exclusive. In fact, in the geneplore model conceptual combinations and analogies are considered as one way to explore new ideas (Ward & Kolomyts, 2010).

**Transformation:** Transformational creativity describes the process of transforming the conceptual space of a problem. This change of the conceptual space can be achieved by "altering or removing one (or more) of its dimensions or by adding a new one" (Boden et al., 1998). Wiggins et al. (2006) further differentiate between transformations of the search space, which we call Search Space Transformation, and transformations of the boundary of the conceptual space, which we denote as Boundary Transformation. More fundamental changes to the conceptual space, like the change of several dimensions or bigger changes in one dimension, lead to the possibility of more varying ideas and, thus, have a higher potential for creative outputs (Boden et al., 1998). Therefore, boundary transformations have a higher potential to lead to a paradigm shift (Wiggins et al., 2006).

Based on our definition, a creative solution has to be both novel and valuable. We introduce two related dimensions to analyze how these two requirements can be met by existing DL systems. One emphasizes covering the entire space (diversity) and the other moving towards the best, locally optimal solution.

**Divergence** is based on the concept of divergent thinking, which describes the ability to find multiple different solutions to a problem (Cropley et al., 2006). Divergence increases the chance of finding more diverse and thus novel solutions.

On the other hand, **convergence** is concerned with finding one ideal solution and is based on the concept of convergent thinking (Cropley et al., 2006). Convergence increases the value of the solution. We apply these two dimensions to the categories based on (Boden et al., 1998).

Figure 3c) visualizes how convergent exploration is guided towards a local optimum, while divergent exploration covers a wider search area, potentially leading towards the global optimum. Figures 3d) and e) visualize convergent and divergent search space and boundary transformation.

In the following chapters, we will discuss how and to what extent these different types of creativity have been achieved in generative deep learning systems.

# 4 FINDINGS

The findings based on our literature review indicate that generative DL is a valuable tool for enabling creativity. DL is aligned with basic processes proposed in models of human creativity. However, while human and AI creativity depends on problem understanding and representation, contextual understanding is far more limited in current DL systems (Marcus et al., 2018). The network is constrained by its training data, lacking the ability to leverage associations or analogies related to concepts not contained in the data itself. The boundary is much more narrow for DL systems than for humans.

**Combination:** The most common way to combine the latent representation of two objects is by using autoencoders. In this case, the latent low dimensional representation of two known objects is combined by vector addition or interpolation. This new latent vector has to be fed back into the decoder network to generate a novel object. An example of this is (Bidgoli & Veloso, 2018), where an autoencoder is trained to learn a latent representation to encode 3D point clouds of a chair. A user can then combine two chairs by interpolating between their latent representations. Several cases in molecule design are also based on autoencoders (Gómez-Bombarelli et al., 2018; Kusner et al., 2017; Polykovskiy et al., 2018). These types of combinations only achieve convergence as they only generate one combination of the two objects. Divergence can be achieved by changing the degree to which the latent dimensions of each input vector contribute to the combined. Human operators can manually control the former.

Combining a trained representation with an unknown input is mostly used in recurrent networks. Here the network is trained to predict the next element in a sequence. This method is mostly used in the language and music domain. Thus, a sequence often consists of letters, words, or musical notes. By providing a new initial sequence for the network to base its prediction on, the contents of this sequence are combined with the representation the network has learned of its training set. One example of this is (Mathewson & Mirowski, 2017), where human actors give the input for a network trained on dialogues. Another prominent example is botnik, a comedy

| Combination | Exploration | Search Space Transformation | Boundary Transformation | Divergence | Convergence | Application Domain | Network Architecture | Paper |
|---|---|---|---|---|---|---|---|---|
| X | | X | X | | X | Image | GAN[1] | (Hu et al. 2019) |
| | X | X | X | X | X | Design | GAN | (Zhibo Liu et al. 2019) |
| | X | X | | X | X | Design | CNN[2], AE[3] | (Bidgoli und Veloso 2018) |
| X | X | X | | X | X | Design | RNN, EV[4] | (Wolfe et al. 2019) |
| X | | X | X | | X | Image | GAN | (Radhakrishnan et al. 2018) |
| X | X | | | X | X | Design | CNN, EV | (Lehman et al. 2016) |
| X | X | X | | | X | Image | CNN | (Gatys et al. 2016) |
| X | X | X | | | X | Image | CNN | (DiPaola und McCaig 2016) |
| | X | | | X | | Image | CNN | (Mordvintsev et al. 2015) |
| | X | | | X | | Image | CNN | (DiPaola et al. 2018) |
| | X | X | X | X | X | Image | GAN | (Elgammal et al. 2017) |
| X | | X | X | | X | Image | GAN | (Zhang et al. 2018) |
| X | | X | X | | X | Image | GAN | (Zhu et al. 2017) |
| X | | X | X | | X | Image | GAN | (Isola et al. 2016) |
| X | X | | | X | X | Image | CNN | (Yalçın et al. 2020) |
| X | | | | X | | Image | CNN LSTM, K-Means | (Karimi et al. 2020?; Karimi et al. 2019; Pegah Karimi et al. 2019) |
| X | | X | | X | | Image | CNN | (Matthew Guzdial und Mark Riedl 2019) |
| | X | X | X | X | X | Design | GAN | (Sbai et al. 2018) |
| | X | X | | X | X | Music | RNN[5] | (Colombo et al. 2017) |
| X | | | | X | | Literature | RNN | (Dean und Smith 2018) |
| X | | X | | X | X | Literature | RNN | (Potash et al. 2015) |
| | X | | | X | | Language | RNN | (Colton et al. 2016) |
| X | | X | | | X | Language, Logic | FFNN[6], Attention | (Brown et al. 2020; Vaswani et al. 2017) |
| X | X | X | | X | X | Humor | RNN | (Mathewson und Mirowski 2017; Piotr Mirowski und Kory Wallace Mathewson 2019) |
| | X | X | | X | X | Science | AE | (Gómez-Bombarelli et al. 2018) |
| | X | X | | X | X | Science | AE | (Kusner et al. 2017) |
| | X | X | X | X | X | Science | AE, GAN | (Polykovskiy et al. 2018) |
| | X | X | X | X | X | Science | RNN | (Segler et al. 2017) |
| | X | X | X | X | X | Science | RNN | (Olivecrona et al. 2017) |
| | X | X | X | X | X | Games | Q-Learning, CNN | (Mnih et al. 2015) |
| | X | X | X | X | X | Games | FFNN[6] MCTS[7] | (Silver et al. 2017) |

1 GAN: Generative Adversarial Network
2 CNN: Convolutional Neural Network
3 AE: Autoencoder
4 EV: Evolutionary Algorithm
5 RNN: Recurrent Neural Network
6 FFNN: Feed Forward Neural Network
7 Monte Carlo Tree Search

Table 1: Concept matrix of the reviewed literature

writing support system, which uses a sequence network to learn a specific writing style. The system then combines this style with text input provided by human operators, generating new texts in the provided style. While this technique converges towards texts of the trained style, interfaces that let human operators choose between the most likely next elements of the sequence can introduce divergence to the process. Another way to use recurrent networks for combinational creativity is to use the entire network for encoding sequential objects. For example, Wolfe et al. (2019) use this technique to encode sequences of gears as recurrent neural networks. By recombining the parameters of different networks, they generate novel sequences. A more complex type of combination is achieved by using style transfer (Gatys et al., 2016). Here a network is trained in a way that allows it to contain a separate representation for the style and the content of an image. These separate representations can be used to combine the content of one image with the style of another one. The most common application of these networks is to combine the contents of photographs with the style of paintings to generate paining like images of a real-world scene (DiPaola & McCaig, 2016). Similar architectures as for style transfer have also been used for numerous other problems, e.g., for unsupervised domain adaptation and even domain prediction (Schneider, 2021). In this case, a DL network might learn to generate samples by identifying and relating concepts from different domains evolving and anticipating their future evolution.

Both autoencoders (Bidgoli & Veloso, 2018) and recurrent networks (Wolfe et al., 2019) can be used to achieve conceptual combinations within a narrow domain, i.e., characteristics (or features) found in the training data. Combinations across domains, i.e., from two (very) different training datasets, were only done using style transfer networks (Gatys et al., 2016). However, these are still restricted to similar domains (e.g., photographs and paintings). This shows that combinational creativity in DL is limited to similar concepts and domains, while humans can form analogies between more different domains.

While many of these instances are limited to combinations of objects in the same or familiar domains, style transfer is an example of combining two different frames of reference as proposed by conceptual combination theory (Ward & Kolomyts, 2010).

**Exploration:** In generative neural networks, explorational creativity can be achieved by searching for new elements in the latent representation learned by the network. The most common way this exploration is implemented in deep learning systems is by introducing an element of randomness. For autoencoders, random samples from the learned latent distribution are fed into the decoder network. Generative Adversarial Networks (GANs) usually use the same process by sampling from the input distribution to the generator network. For sequential data, recurrent neural networks (RNNs) can be trained to predict the next elements in a sequence. Using randomly initialized sequences, new sequences can be generated (Graves et al., 2013). The initial element of a sequence is randomly generated and used to predict the most likely consecutive elements under the data representation learned by the model. This sampling process from a latent space can be interpreted as an instance of random search (Solis and Wets 1981). However, instead of searching the problem space, the lower-dimensional representation learned by the network is searched. Due to the use of random search, these methods do not converge towards an optimal output and can only ensure divergence.

Convergence can be added to the exploration of the search space by applying more complex search algorithms. Examples are using gradient search (Bidgoli & Veloso, 2018) or even reinforcement learning (Olivecrona et al., 2017). A special case of exploration that takes the novelty of the generated example into account is the application of evolutionary algorithms in combination with neural networks. This shows that, while most baseline instances of explorational creativity in DL are limited to simple random search processes, more complex search strategies are possible in the search space defined by the network's features. Thus, the extent of creativity achieved via exploration is mostly limited by transformational creativity.

**Search Space Transformation:** Autoencoders are initially trained to learn a latent data representation. The decoder ensures that the reconstructions from this latent space belong to the same distribution as the training data, thus ensuring convergence towards the training data set while leaving divergence to the exploration of the trained latent representation. For sequential data, recurrent neural networks (RNNs) can be trained to predict the next elements in a sequence, thus enabling a convergent search space transformation (Graves et al., 2013).

Generative Adversarial Networks (GANs) are trained to generate outputs from the same distribution as the training data out of random inputs. In addition to the generator network, a discriminator network is trained to differentiate the generator's output from real data. In this way, the performance of the discriminator improves the quality of the generators' outputs (Goodfellow et al., 2014). In contrast to autoencoders, GANs already contain divergent processes in the training phase. Already during training, the generator is passed randomly sampled inputs, adding a divergent element to the parameter training. The convergence of these outputs is achieved by training the generator to produce outputs indistinguishable from the training data. Still, it is very difficult for GANs to produce realistic diverse images such as natural images. According to Wang et al. (2021), achieving this "Mode Diversity" is one of the most challenging problems for GANs. SAGAN and BigGAN address this issue with specific model architectures, while SAGAN and BigGAN apply

CNNs with self-attention mechanisms to increase diversity.

Elgammal et al. (2017) make use of theories on creativity to extend GANs to creative GANs. They added network loss, penalizing outputs that fit well into a known class structure expected to encode different styles. By optimizing the GAN to generate outputs with a high likelihood perceived as art but with a low likelihood fit in any given artistic style, they aim to optimize the arousal potential of the resulting image for human observers.

In reinforcement learning, where an agent interacts with the environment based on rewards, exploration is explicitly encoded in the agents' behavior. This is done to prevent the agent from learning suboptimal strategies due to limited knowledge of the environment (Sutton & Barto, 2011). In reinforcement learning, the interaction between convergence and divergence can be seen as equivalent to the tradeoff between exploration and exploitation.

We can see that convergent search space transformation is achieved in almost all examples by the standard training mechanisms of neural networks. To achieve divergence, more complex architectures or loss regularizations are required. However, in most cases, convergence is limited to ensuring similarity with the training data. The only example we could find that actively trained a network towards novelty of the outputs and can therefore be considered as divergent search space transformation was achieved using an alternative training mechanism for neural networks based on evolutionary algorithms (Wolfe et al., 2019).

**Boundary Transformation:** Honing theory describes a recursive process in which the problem domain is reconsidered through creation, which in turn is based on the current understanding of the problem domain (Gabora et al., 2017). In GANs the interaction of the generator and the discriminator can be interpreted in the same way. The understanding of the problem domain is given by the discriminator's ability to decide between a true and a fake object. The generator's goal is always to generate realistic objects under the model of the problem domain. By using feedback of the discriminator based on the generated objects, the domain model, i.e., the generator, is altered. In deep reinforcement learning, a similar effect can be observed as the loss of the policy or value network changes with discovering additional states and rewards. However, on a higher level, the overall task of the network still stays the same, whether it is generating realistic outputs for GANs or maximizing the rewards for reinforcement learning. Segler et al. (2017) introduce a mechanism similar to Honing to the task of sequence learning. They first train an RNN to generate molecule sequences using a large and general training set of molecules for training. They then use an additional classification system to filter all highly likely molecules to show a required attribute from all randomly generated examples. For Honing their generator, they fine-tune the RNN only on this set of selected molecules. This process is iteratively repeated several times.

However, as these mechanisms only impact the training mechanism by generating new training data, they can only impact one aspect of the boundary. Additionally, both these mechanisms only transform the boundary in a convergent fashion. They further restrict the conceptual space towards containing valuable solutions at the cost of novelty. More complex boundary transformations still require either a human operator's choices or can be achieved through meta-learning.

**Convergence/Divergence:** Divergence in a given search space relies heavily on random inputs. While there are complex methods to achieve convergence in a given search space (Olivecrona et al., 2017), few applications use them. Transformation of the search space is mostly limited to convergence. This holds even more for transformations of the boundary. DL techniques do not enforce divergent transformations. While this might be achieved by adding regularization terms to the training loss, divergent boundary transformations seem harder to achieve in contemporary DL models.

# 5 DISCUSSION AND FUTURE WORK

The findings based on our literature review indicate that generative DL is a valuable tool for enabling creativity. DL is aligned with basic processes proposed in models of human creativity. However, while human and AI creativity depends on problem understanding and representation, contextual understanding is far more limited in current DL systems (Marcus et al., 2018). That is, the network is constrained by its training data, lacking the ability to leverage associations or analogies related to concepts that are not contained in the data itself. The boundary is much more narrow for DL systems than for humans. DL techniques do not enforce divergent transformations. While this might be achieved by adding regularization terms to the training loss, divergent boundary transformations seem harder to achieve in contemporary DL models.

So far, all these transformations are limited to small incremental changes in the representation and are heavily dependent on the training data. More fundamental changes, that take other domains into account are still left to the humans designing the models, as can be seen in the decision to use a text like representation for complex three-dimensional molecule structures, which allowed the use of models previously successful in text generation (Segler et al., 2017).

Many domains highly depend on human creativity. They either completely lack large amounts of data for training generative DL systems or a creative solution might rely on characteristics that rarely occur in the data. This means that the results are highly dependent on the quality and even more so the quantity of the training data. This can also be seen by the fact that creative applications are mostly found in domains, where DL already performs well on non-creative tasks, like images (e.g., Gatys et al., 2016) or (short) texts (e.g., Dean & Smith, 2018). At the same time, it is still an open problem for a DL system to generate long continuous texts that tell a coherent, novel story, just as it is a hard problem, to automatically summarize longer stories and answer complex questions that require contextual knowledge.

Concerning the level of creativity in the observed literature, most models can produce only everyday creativity (little-c). One could argue that the examples of de-novo drug design constitute an example of pro-c creativity. However, because the final selection and synthetization of the promising molecules still require human experts, they merely support pro-c creativity. The only example that could be argued to possess Big-C creativity is Alpha-Go (Silver et al., 2017). It achieved the level of a world champion in its domain and could generate strategies that humanize expert players later adopted. A creative capability that is currently beyond AI, is the ability to identify the existence of a problem or the lack of creative solutions in the first place. Thus, creative AI is still far from the capabilities covered by problem-finding theories of creativity.

While our findings indicate that the creativity of DL is highly limited, DL has a key advantage compared to humans: It can process large amounts of data. Given that DL systems are currently trained on very narrow domains, their creative capabilities might increase merely because of more computational power, allowing them to explore a larger space of possible creative solutions than today. Furthermore, many DL systems are simple feedforward networks. Advances in reasoning of neural networks, such as reflective networks (Schneider and Vlachos, 2020), could also enhance creativity. Even more, meta-learning might adjust the boundary, which is not commonly done in existing work. However, even given that more training data and meta-learning are used, human creativity is likely not reached: Humans must define the framework for meta-learning. In the end, they must be creative in the first place to derive new methods, also allowing for longer chains of reasoning and models that allow for more sophisticated transformations of the conceptual space.

In future research, we plan to compare the creative capabilities of DL with computational creativity systems based on other models like evolutionary algorithms and cognitive models. Not only can this help to compare the capabilities of different models, but it might also lead to new ways to improve DL's creative capabilities by adapting concepts from other models. We also want to study applications of human-AI interaction for creative tasks and enhance our conceptualization accordingly

# 6 CONCLUSION

Deep learning shows a large potential for enabling the automation and assistance of creative tasks. By linking the functionality of generative deep learning models with theories of human creativity, we provide an initial step in better understanding the creative capabilities of these systems and the shortcomings of current models. Our analysis showed that deep learning possesses many traits of computational creativity, such as combinatorial or Darwinian exploration, but novelty is strongly constraint by the training data. We hope that this knowledge helps practitioners and researchers to design even better systems for supporting humans in performing creative tasks and to assess the suitability of deep learning for creative applications for businesses and the social good.